\documentclass{article}

\usepackage{microtype}
\usepackage{multirow}
\usepackage{graphicx}
\usepackage{booktabs} 

\usepackage{hyperref}
\usepackage{subcaption}

\usepackage[accepted]{icml2019}

\icmltitlerunning{An Empirical Analysis of the Impact of Data Augmentation on Knowledge Distillation}

\begin{document}

\twocolumn[
\icmltitle{An Empirical Analysis of the Impact of Data Augmentation on Knowledge Distillation}



\icmlsetsymbol{equal}{*}

\begin{icmlauthorlist}
	\icmlauthor{Deepan Das}{equal,to}
	\icmlauthor{Haley Massa}{equal,to}
	\icmlauthor{Abhimanyu Kulkarni}{equal,to}
	\icmlauthor{Theodoros Rekatsinas}{equal,goo}
	
\end{icmlauthorlist}

\icmlaffiliation{to}{Department of Electrical and Computer Engineering, University of Wisconsin-Madison, USA}
\icmlaffiliation{goo}{Department of Computer Science and Statistics, University of Wisconsin-Madison, USA}


\icmlcorrespondingauthor{}{ddas27@wisc.edu, hmassa@wisc.edu}


\vskip 0.3in]



\printAffiliationsAndNotice{}  

\begin{abstract}

Generalization Performance of Deep Learning models trained using Empirical Risk Minimization can be improved significantly by using Data Augmentation strategies such as simple transformations, or using Mixed Samples. We attempt to empirically analyze the impact of such strategies on the transfer of generalization between teacher and student models in a distillation setup. We observe that if a teacher is trained using any of the mixed sample augmentation strategies, such as MixUp or CutMix, the student model distilled from it is impaired in its generalization capabilities. We hypothesize that such strategies limit a model's capability to learn example-specific features, leading to a loss in quality of the supervision signal during distillation. We present a novel Class-Discrimination metric to quantitatively measure this dichotomy in performance and link it to the discriminative capacity induced by the different strategies on a network's latent space.
\end{abstract}

\section{Introduction}
\label{submission}
\begin{figure}[ht]
	\vskip 0.2in
	\begin{center}
		\centerline{\includegraphics[width=\columnwidth]{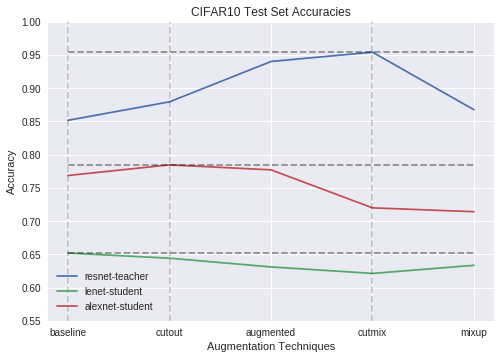}}
		\caption{Overview of the results of Mix Up, Cut Mix, Cut Out and simple transformations on CIFAR-10 Test Set when distilled into two different student models from a ResNet Teacher Model. Note that Mixed Sample Augmentation strategies help improve teacher performance, but corresponding student performance is impaired.}
		\label{fig:overall}
	\end{center}
	\vskip -0.4in
\end{figure}

Deep Neural Networks (DNN) now have the ability to generate sophisticated models that capture the intricacies of information in large amounts of data. This ability has lent spectacular success to Deep Learning models in many domains. However, often as the ability for the machine learning models to encapsulate information needed to express this complexity increases, so does the size of the machine learning model. These complex models may be too demanding to run on more reasonable hardware, including personal computers or mobile devices. One solution to this problem is to use knowledge-based distillation, which trains a smaller, more efficient model that approximates the performance of the original, more cumbersome model. A knowledge distillation objective can enable us to train a smaller, lightweight model without using any external annotations on given data, but instead utilising the prediction labels generated by a pre-trained cumbersome teacher model. There are plenty of other techniques that achieve high performance in compressed models including model quantization \cite{zhou2017convergence}, model pruning \cite{han2015deep}, and more recently, lottery tickets \cite{frankle2018lottery}.

 \begin{figure*}[ht]
	\vskip 0.2in
	\begin{center}
		\centerline{\includegraphics[width=1.95\columnwidth]{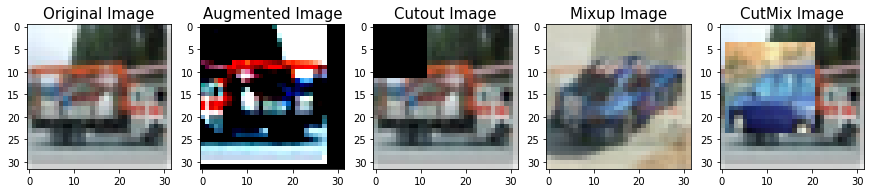}}
		\caption{Pictorial representation of the data samples generated using the several data augmentation strategies. Starting from the left, we note that the original image is that of a firetruck. Standard transformations such as flipping and cropping produces the augmented image. Cutout introduces regional dropout in the image, whereas in Mix Up and Cut Mix, images and their labels are combined proportionally. For instance, here we see that a car is combined along with the firetruck.}
		\label{fig:intro-image}
	\end{center}
	\vskip -0.2in
\end{figure*}

 With the recent surge of deep learning research,  there is a natural need to create knowledge-based distillation techniques that mimic more complex algorithms. At the same time, however, there is little basis for evaluating the impact of existing strategies on the latent representation space generated by the teacher model that allows distillation training on the student model. This raises the question: what is the impact of generalization in teacher networks? While many forms of explicit regularization have been added to large, cumbersome models, it will be interesting to analyze the impact of the implicit regularization techniques when applied to such neural networks through the different optimization techniques. Implicit regularization has shown that as neural networks increase in size, they are actually able to create solutions with lower complexity \cite{neyshabur2017implicit}. Since there is no explicit regularization in the model’s objective function, this ability to generalize is built in through well studied techniques such as normalization, gradient descent, and the choice of weight kernels, like the convolution filter. Some of these regularizers have become fairly popular and have become a standard addition to any Deep Learning model optimization.

Regularization performed on neural networks have many different aspects. In a distillation setup, where a smaller model(student) attempts to mimic the softened softmax output of a large, cumbersome model(teacher), student performance is shown to improve, but there still exists a significant generalization gap between the two models. It is indeed important to try and minimize this generalization gap, but it is also important to consider the viability of using one or many of these implicit regularizers in the distillation process. In this experiment, we will be creating a framework  that measures the impact of using such implicit regularizers on the generalization gap between teacher and student models. 

One such form of implicit regularization is the Vicinal Risk Minimization (VRM) principle \cite{vapnik2000vicinal}. In this work, we will evaluate the impact data augmentation inspired by VRM techniques have in the transfer of generalization between teacher and student models. However, this is a general framework, and can also be used to evaluate other types of generalization in the future. In any natural setting, we encounter noise in the presented data that enables us to gain additional information about relatively similar data. Traditionally, data augmentation and even Cut Out \cite{devries2017improved}, is viewed as one such example, where the model is trained on some simple transformations of the data. However, the scope of a vicinity is not explicitly defined in this case. Mixed Sample Data Augmentation Techniques like Mix Up  \cite{zhang2017mixup} and Cut Mix \cite{Yun_2019} or FMix \cite{harris2020understanding} provide a new outlook on the concept of vicinity. Considered as standalone techniques, they can lead to state of the art results in standard Deep Learning tasks, but it will be interesting to note the impact each technique has on the supervision signal from the teacher model to the student model. We hypothesize that even though Data Augmentation techniques provide good regularization, they impair the distillation process because of several implicit qualitative biases in the techniques. This impairment is much more pronounced in the Mixed Sample Data Augmentation Techniques.

To prove this hypothesis, we will be using a framework to evaluate four data augmentation techniques inspired by VRM (traditional augmentation, Mix Up, Cut Out and Cut Mix) by creating teacher networks on ResNet-18 architectures. Student networks will then be distilled and evaluated against appropriate data sets to investigate their generalization capabilities. Our contributions are thus summarized as follows:

\begin{itemize}
	\item We demonstrate that popular data augmentation techniques, and especially Mixed Sample techniques, such as MixUp and CutMix when applied on a teacher model, can impair the transfer of generalization capabilities onto a student model in a distillation setting. 

\item We present a novel similarity-based metric to help explain some qualitative traits inherent in the latent representations of such models. These findings are also backed by a traditional KL-Divergence based metric that operates on the probability distribution of the model predictions.

\item We also analyze the performance of these distilled models under distributional shift, and demonstrate the adversarial impact of Mixed Sample Augmentation strategies on the distillation objective. 

\item We present empirical proof that data augmentation techniques tend to increasingly make models more discriminative and regularizes on example-specific features pertinent to the image. 
\end{itemize}

\section{Related Work}

\textbf{Knowledge Distillation}: The idea behind knowledge distillation was first introduced in 2006 \cite{bucilua2006model}, but made popular by \cite{hinton2015distilling} in 2015, as a novel technique for model compression. It works by teaching a student network to mimic, step by step, the behavior of a larger network. This process works by first obtaining a more cumbersome teacher model on the original data. The complex features thus extracted by the teacher model are then transferred to a simpler model or a model of similar size \cite{furlanello2018born} by using softened probability scores from the cumbersome models. It is important to note that the relative probability of incorrect cases to the probability of the correct class and each other are useful in understanding how complex models generalize the data. For example, in the well-known MNIST dataset, it is helpful to understand not only that the correct classification is a 2, but also which of these 2’s look like 3’s and the 2’s that look like 7’s. Based on this idea, several improved distillation techniques have been proposed \cite{tarvainen2017mean, Li_2017, xie2019selftraining, Heo_2019} that address different aspects of improving distillation performance. More recent work attempts to find the crucial aspects that determine the quality of a distilled model \cite{phuong2019towards}.

\textbf{Regularization}: The ability of the cumbersome model to learn the nuances in probability can be seen as the model’s ability to generalize the data. Generalization, can be interpreted as the capacity of a model to adapt to new, unseen data, drawn from the same distribution the model was trained on. Several works analyse the generalization performance of the numerous implicit regularization methods in Deep Learning. Moreover, several other papers propose different techniques of introducing noise to improve generalization performance of highly parameterized neural networks, \cite{Srivastava2014DropoutAS, blundell2015weight, ioffe2015batch, vapnik2000vicinal}. Some other works also analyse the relationship between certain qualities of a learned model and the measured performance metrics \cite{neyshabur2017implicit, tsipras2018robustness, Peterson_2019}. Knowledge distillation as a technique for model compression is unique due to its similarity to human learning, and is thus an interesting downstream task to analyze more about a neural networks ability to generalize.  A lot of research has been done to improve the generalization of the student by re-defining a novel teacher architecture, such as adding sequence-level techniques and ensemble teachers \cite{Simard1996TransformationII}. However, this work is more interesting in exploring the impact of generalization techniques meant to enhance the existing teacher architecture. Some recent works attempt to explain the relationship between implicit regularization techniques and knowledge distillation. \cite{mller2019does} used a similar approach to ours while using a novel latent representation strategy to model the adversarial impact of label smoothing on distillation, while \cite{arani2019improving} analyse the beneficial impact of using trial-to-trial variability during distillation. \cite{Cho2019} analyse the impact of Early Stopping on Distillation, but none refer to data augmentation strategies. Specific to Data Augmentation techniques, there have been attempts to explain the latent effect of data augmentation using mathematical formulations, such as in \cite{chen2019grouptheoretic, thulasidasan2019mixup, he2019data}, and for Mixed Sample Data Augmentation Techniques in \cite{harris2020understanding}. In the following sections, we describe the experimental setup used in this paper, and then attempt to explain the interesting results obtained. 

\section{Experimental Setup}

In this section, we will describe in detail the experimental setup used in this paper, ranging from the VRM techniques considered for the experiments, to the generalization measures we used to evaluate the student and teacher networks.

\subsection{Comparison Methods}

The principle behind VRM was introduced in \cite{vapnik2000vicinal}, and has found its application in Machine Learning model training via several different tools. Using the standard Empirical Risk Minimization principle, the loss objective is optimized only on the training samples, whereas in VRM, virtual data points are also sampled from the vicinity of the real data points. Thus, whereas ERM can be thought of as minimizing the expectation of the loss objective with respect to an available empirical distribution $P_{emp}(\mathbf{x}, y)$, VRM can be thought of as a natural improvement where the density estimates on each sample is replaced by some estimate of the density in the neighborhood of the sample. Thus, the optimization objective is now regularized by an uncertainty estimate in the sampling task as follows:
\begin{center}
    $$ R_{vicinal}(f) = \frac{1}{n} \sum_{i=1}^{n}  \int l(f(\mathbf{x}), y) dP_{x_i}(x) $$
\end{center}

The following augmentation strategies can be thought of as extended VRM techniques and are used in the experiments to follow. These strategies are represented pictorially in \ref{fig:intro-image}.

\begin{table*}[htbp]
	\caption{Performance of the different Teacher and Student Models on CIFAR-10 Test Set. The KLD Metric is the distance between Human labeled confidence scores and Model prediction probabilities. Expected Calibration Error(ECE) measures prediction quality}
	\label{table:cifar-metrics}
	\vskip -0.3in
	\begin{center}
		\begin{sc}
			\begin{tabular}{lccccccccc}
				\toprule
				& \multicolumn{3}{c}{Teacher} & \multicolumn{3}{c}{LeNet Student} & \multicolumn{3}{c}{AlexNet Student} \\
				Models & Accuracy & KLD & ECE  & Accuracy & KLD & ECE & Accuracy  & KLD  & ECE\\
				\midrule
				Baseline & 0.852 & 0.656 & 0.205 & \textbf{0.652} & 1.002 & 0.15 & 0.769 & 0.710 & 0.149\\
				Augment & 0.940 & 0.466 & 0.255 & 0.631 & 0.951 & 0.070 & 0.777 & 0.735 & 0.210\\
				Cutout & 0.880 & 0.220 & 0.237 & 0.644 & 0.987 & 0.109 & \textbf{0.785} & 0.726 & 0.249\\
				Mixup & 0.868 & 0.641 & 0.170 & 0.633 & 0.991 & 0.033 & 0.714 & 0.836 & 0.130\\
				Cutmix & \textbf{0.954} & 0.524 & 0.252 & 0.621 & 0.987 & 0.28 & 0.720 & 0.776 & 0.062\\
				\bottomrule
			\end{tabular}
		\end{sc}
	\end{center}
\end{table*}

 \begin{table}[htbp]
	\caption{Performance on CINIC-Imagenet Test Set.}
	\label{table:cinic-metrics}
	\begin{center}
		\begin{sc}
			\resizebox{\columnwidth}{!}{\begin{tabular}{cccc}
					\toprule
					
					Models & Teacher & LeNet-Accuracy & AlexNet-Accuracy \\
					\midrule
					Baseline & 0.600 & \textbf{0.457} & 0.544 \\
					Augment & 0.687 &  0.439 & 0.549 \\
					Cutout & 0.630 & 0.451 & \textbf{0.555}  \\
					Mixup & 0.614 & 0.444 & 0.498  \\
					Cutmix & \textbf{0.716} & 0.439 & 0.499 \\
					\bottomrule
			\end{tabular}}
		\end{sc}
	\end{center}
\end{table}

\textbf{Standard Transformations}: Data Augmentation techniques such as flipping, splitting, scaling, rotating, cropping, etc. are some of the most common techniques to augment data for image classification techniques and find their use in several state of the art results. While general data augmentation leads to improvement in generalization by creating several invariant features, it is important to note that these augmentations are data set dependent and require domain expertise. In this work, a few traditional data augmentation techniques are tested, that include random cropping and random flipping along the horizontal axis.

\textbf{Cut Out}: Cut Out is a generalization technique inspired by the Dropout  regularization technique \cite{dropout2014}. While Dropout has been shown to improve generalization of the model, it blindly reduces the capacity of the model by adding in sensitive hyper-parameters \cite{Konig2019}. Therefore, Cut Out works as a data augmentation technique, pulling from previous works that show that dropping continuous areas has shown improvement in generalization \cite{DropBlock2018}. A selected number of randomly sized continuous sections are removed from the image to create a modified image for training, as given by \cite{devries2017improved}.

\textbf{Mix Up}: A relatively newer strategy to implement data augmentation is to create a convex combination of data samples and their label. The idea was first introduced in \cite{zhang2017mixup} and the virtual data-label pair can be generated as follows: 

\begin{center}
    $\hat{x} =  \lambda x_i + (1 - \lambda) x_j$ \\
    $\hat{y} =  \lambda y_i + (1 - \lambda) y_j$ 
\end{center}

where $\lambda \sim \beta(\alpha,\alpha)$ is the mixture percentage of each image. This creates a target value that is a mix of the two original target values. By using an implicit bias that linear interpolations of data should lead to predictions that are linearly interpolated in the target space, Mix Up enables generation of well-calibrated models whose generalization performance is slightly better \cite{thulasidasan2019mixup}.

\textbf{Cut Mix}: Introduced by \cite{Yun_2019}, Cut Mix was inspired by augmentation techniques that blend the classes of images \cite{Tokozume2018}, like Mix Up, and techniques that cut out regions of images, like Cut Out. The new virtual sample is given as: 

\begin{center}
    $\hat{x} =  \mathbf{M} \odot  x_i + (1 - \mathbf{M}) \odot  x_j$ \\
    $\hat{y} =  \lambda y_i + (1 - \lambda) y_j$ 
\end{center}

where $\mathbf{M}$ is a binary mask that contains the information of where to drop and fill the image, $\odot$ denotes element-wise multiplication and $\lambda \sim  \beta $(1, 1) is the combination ratio. The target label is generated similarly to Mix Up, but the authors claim that it improves upon both Cut Out and Mix Up by not removing informative pixels and generalizing over more natural samples.

We understand that none of these techniques can be truly considered to be pure VRM techniques in the absolute sense, but only capture the essence of VRM. However, these techniques are some of the most popular regularization techniques to be found in the state of the art results for most Deep Learning tasks. Moreover, a lot of recent work attempts to understand the qualitative abilities of such techniques \cite{he2019data, gontijolopes2020affinity}

\subsection{Knowledge Distillation}

As described above, we attempt to understand the impact of these augmentation strategies in a simple Knowledge distillation setup, where a smaller model is trained by forcing it to mimic the tempered probability distributions of a larger, cumbersome model. When neural networks use a softmax output layer to convert the logit $z_i$ for an example $x_i$ into a probability score $p_i$, we can soften it using a temperature parameter $T$ in the following manner: 

\begin{center}
	\[p_i = \frac{exp(z_i / T)}{\sum_{j}^{N} exp(z_j / T)} \]
\end{center}

One can conveniently use unlabeled data to train a smaller model, by completely depending on the cumbersome model's softened outputs, or use label information, if available while using the following loss formulation using $\tau$ as the temperature parameter.

\begin{center}
	\[\mathcal{L}(\mathbf{x}, y) = (1 - \alpha) \mathcal{L}_{CE}(f_S(\mathbf{x}), y) + \alpha \tau^2 \mathcal{D}_{KL}(f_S^{\tau}(\mathbf{x}), f_T^{\tau}(\mathbf{x}))\]
\end{center}

\subsection{Evaluation}

After having trained a set of teacher and student models using the above mentioned augmentation and distillation strategy, it is important to set up a well-defined and intuitive evaluation strategy that can explain the impairment in transfer of distilled knowledge. 

\subsubsection{Test Datasets} A key aspect of measuring generalization performance is to analyse performance of the models on not only unseen data, but also test data with some natural variations in it. For this, we use the following data sets to analyse the performance of the models trained on CIFAR-10.

\textbf{CIFAR-10}: This is used to measure the performance of the model on unseen data lying within the seen distribution. 

\textbf{CINIC-10}: This is used for the out-of-sample generalization test. This data set is collected by \cite{darlow2018cinic10} and contains both CIFAR-10 and ImageNet images in its test fold. However, we just use the 70,000 ImageNet images that have been bucketized into CIFAR-10 classes. 

\begin{figure}[ht]
	\vskip 0.2in
	\begin{center}
		\centerline{\includegraphics[width=\columnwidth]{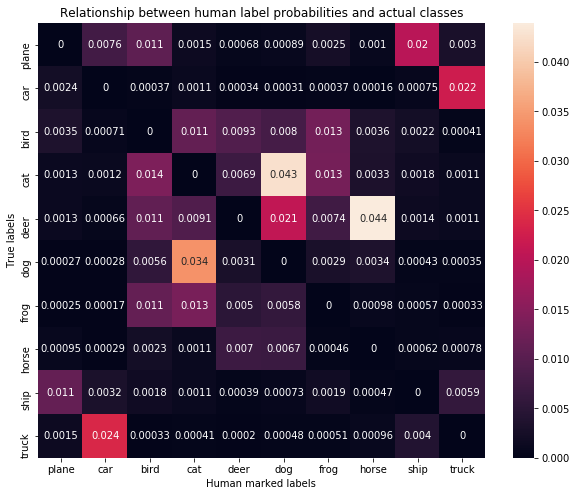}}
		\caption{Average human confidence distribution across ground truth classes in CIFAR-10. Diagonal elements, that have the highest confidence have been masked to reveal implicit patterns between different classes}
		\label{fig:human-dist}
	\end{center}
	\vskip -0.2in
\end{figure}

\textbf{CIFAR-10H}: This is called the Human Labeled CIFAR Test Set which is collected by \cite{Peterson_2019} and contains the exact same images as the CIFAR-10 images, but instead of one-hot encoded target labels, this dataset provides the original probability distributions based on the labelings provided by the human annotators. This enables us to measure the closeness of a model's predictions with human beliefs about a data sample.

\subsection{Generalization Measures}
We evaluate each model on the basis of standard confusion metrics, such as \textit{Accuracy, F-1 Score, Precision and Recall}. However, accuracy ignores the probability assigned to the predicted class label as well as the probabilities assigned to the incorrect classes. This information is key to understanding the degree of generalization that a model has achieved. Since we have access to the probability distributions for each CIFAR test sample, we can easily compare a model's softmax score vector with the human-labeled distribution. To better understand this, note that in Figure \ref{fig:human-dist}, which represents the average confidence scores that humans have in each class, certain implicit patterns exist, that reveal a more generalized story for a prediction. For instance, humans are more likely to confuse a dog with a cat than a car. This generalization should also exist in the predicted models. This difference in model prediction and human-level confidence can be easily measured by a KL-Divergence between the model prediction and human-labeled scores, as formulated below, where $p_f(x_i)$ represents the model output and $p_h(x_i)$ represents the human labeled confidence scores.

\begin{center}
	\[ \mathcal{L}_{CE}  = \sum_{i=1}^{N} \mathcal{D_{KL}}(\mathbf{p}_f(x_i), \mathbf{p}_h(x_i)) \]
\end{center}

In essence, this is nothing but the \textit{cross-entropy loss} between the model predictions and human labels. However, it would also be interesting to note the KL-Divergences between different classes. Since, it is not feasible to do that on a sample-level, it would be interesting to compute the Divergences between the averaged probability distribution for each ground truth class for both the model prediction scores and human labeled scores. To understand quality of predictions, we also plot reliability diagrams of the different models and compare the model calibration using \textit{Expected Calibration Error}, which is given as follows:
\begin{center}
	\[ ECE =  \sum_{m=1}^{M} \frac{|B_m|}{N} |acc(B_m) - conf(B_m)|  \]
\end{center}
 
 We also propose a novel metric to explain the discriminative power of the different models we train. This operates on the penultimate layer embeddings generated by the model and can be thought of as a measure of how well-separated the embedding manifold is. It takes into account both the intra-class and the inter-class similarity and defines the discriminative power of the model as the difference between them. If intra-class similarity is high, class representations are cohesive and more compressed. If inter-class similarity is low, class representations are less adhesive and are far away from each other. An optimal classifier will tend to have high cohesivity and low adhesion, and thus higher discriminative power. To compute this metric, we first standardize the embeddings, and define the cohesion and adhesion metrics as inter- and intra-class similarities, respectively using a cosine similarity function $S$.
 
 \begin{center}
     \[ C^{(i)} = \frac{1}{N_C(N_C - 1)} \sum_{i=1}^{N_C - 1} \sum_{j=i+1}^{N_C} S(d_i, d_j)\]
     \[A^{(i,j)} = \frac{1}{N_1 N_2} \sum_{i=1}^{N_1} \sum_{j=1}^{N_2} S(d_i, d_j) \]
 \end{center}
 Where, $N_k$ represents the number of instances in Class $k$. Thus, the class-discrimination can then be computed as:
 
 \begin{center}
     \[ D = \frac{1}{\sqrt{d}}[ \frac{1}{K} \sum_{i=1}^{K} C^{(i)} -  \frac{2}{K(K-1)} \sum_{i=1}^{K - 1} \sum_{j=i+1}^{K} A^{(i,j)}]    \]
 \end{center}
 Where, $K$ represents the total number of classes and $d$ represents the dimensionality of the embeddings.

\begin{figure*}[htbp]
	\vskip 0.2in
	\begin{center}
		
		\begin{minipage}{.22\textwidth}
			\centering
			\includegraphics[width=1.1\textwidth]{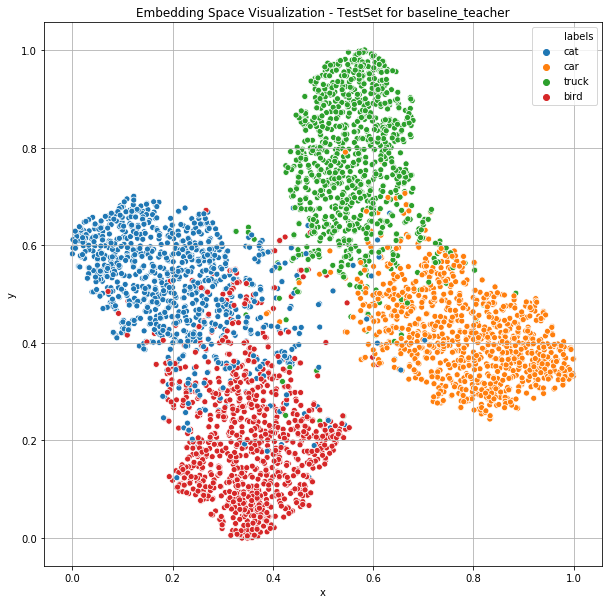}
			\subcaption{Baseline}
			
		\end{minipage}%
		\begin{minipage}{.22\textwidth}
			\centering
			\includegraphics[width=1.1\textwidth]{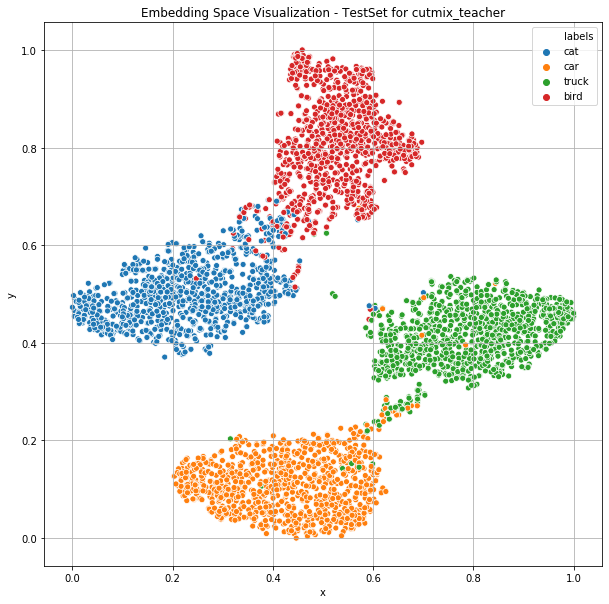}
			\subcaption{CutMix}
		\end{minipage}%
		\begin{minipage}{.22\textwidth}
			\centering
			\includegraphics[width=1.1\textwidth]{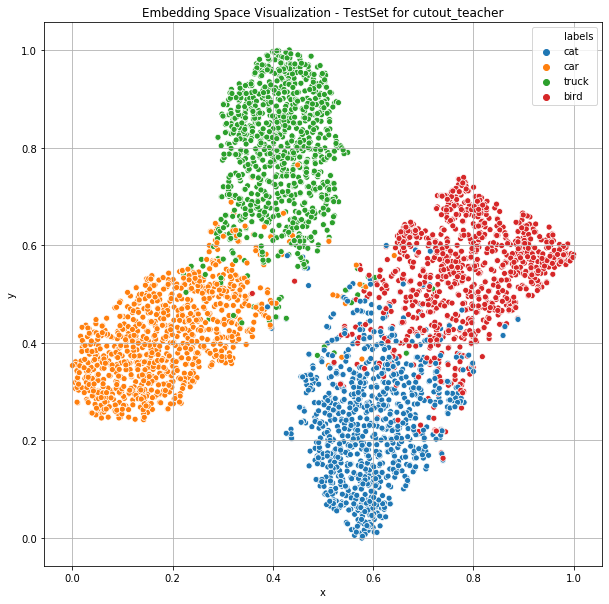}
			\subcaption{Cutout}
		\end{minipage}
		\begin{minipage}{.22\textwidth}
			\centering
			\includegraphics[width=1.1\textwidth]{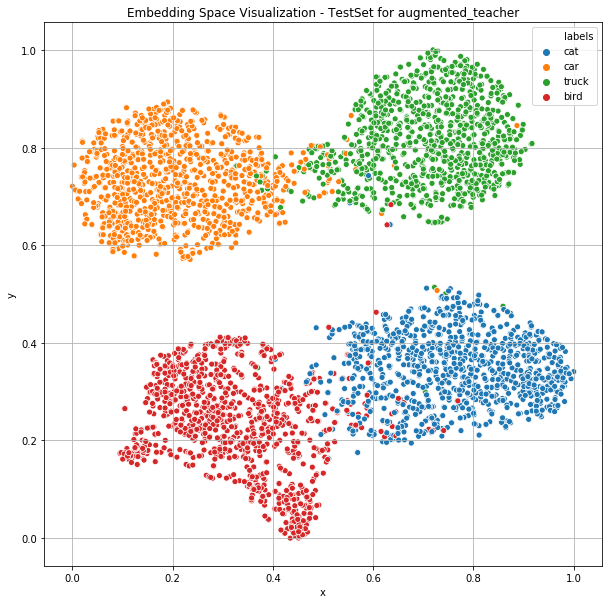}
			\subcaption{Augmentation}
		\end{minipage}
		\caption{Latent Space Visualization from Penultimate Layer, dimensionally reduced using tSNE. Note the well-separated class representations in the augmented models as compared to the baseline model}
		\label{fig:latent-space}
	\end{center}
	\vskip -0.2in
\end{figure*}

\section{Experiments}

Five unique experimental setups were created using the framework and evaluation techniques from above. Teacher models were trained using augmentation techniques and distilled into the student models. This was evaluated with different complexity student models, and an additional data set for comparison. Additionally, we also performed a small ablation study, where we compared the effect of our augmentation techniques on each step of the knowledge distillation process.

The models were implemented in PyTorch, with different parameter values for the different augmentation techniques. Knowledge distillation was performed with a temperature value of 20 and gamma of 0.5. When performing augmentation, baseline and standard data augmentation did not require explicit parameters. Mix Up was implemented with a beta value of 1, and Cut Out was implemented with 16 holes to be randomly places in the images. Finally, Cut Mix used alpha and beta parameters of 0.5 and 1, respectively. These values were chosen based on original paper implementation of the techniques, and could be fine tuned in later work. 

\subsection{CIFAR 10 - Augmented ResNet Distilled to LeNet}

\subsubsection{SetUp}
Five ResNet18 models \cite{He2015} were trained using the data augmentation techniques discussed above, on the CIFAR-10 training dataset. The student networks were built from the LeNet architecture \cite{LeCun1998} with no data augmentation techniques added.

\subsubsection{Results}

As can be seen in \ref{table:cifar-metrics}, the baseline student model out performed other models in terms of accuracy. The baseline model also outperformed all other student models in terms of Precision, Recall and F1 Score on the CIFAR-10 and CINIC data set, which can be found in the appendix \ref{table:student-metrics}. 

\subsection{CIFAR 10 - Augmented ResNet Distilled to AlexNet}

\subsubsection{Set Up}
Another experiment distilled the ResNet18 teacher networks, from above, into a more complex student architecture, to evaluate if it's increased complexity would better capture the nuance of the Teacher network generalization. The architecture chosen was the AlexNet \cite{NIPS2012_4824}. 

\subsubsection{Results}
This experiment emphasized the same trend from the previous experiment that enhanced generalization did not lead to equally enhanced distillation, as can be seen in \ref{table:cifar-metrics}, \ref{table:student-metrics}. However, it is important to note that the augmentation techniques based on assumed sampling distance, such as standard augmentation and Cut Out, outperformed the baseline model. While, augmentation techniques that incorporate VRM sampling, Mix Up and Cut Mix, still under-performed the baseline teacher-student model. It is of interest that possibly the larger AlexNet model was able to pick up on more nuance. Additionally, we propose that this difference between more standard augmentation and Mix Sample augmentation techniques is related to the varying interpolative nature of the techniques. This finding will be explored and further evaluated in the next section.

\subsection{MNIST - Augmented AlexNet Distilled to LeNet}

\subsubsection{Set Up}

This work also used the MNIST data set to validate these findings on another data set. For this validation, we just compared the baseline ERM technique with our Mixed Sample augmentation techniques - Mix Up and Cut Mix. This was due to the fact that the standard augmentation techniques as we defined them have been shown to inhibit MNIST training \cite{he2019data}. We trained three teacher networks using the AlexNet architecture on the MNIST data using the augmentation techniques. Information from these models was then distilled into baseline LeNet models, which were then evaluated. 

\subsubsection{Results}

The results can be seen in the table below. While the accuracies were very close to one another, it is shown that the validation loss shows more significant decrease for the baseline model. This validation was done with fairly simple models and data set, and it would be interesting to further explore this concept with more complex models and data sets for further work. 

\begin{table}[htbp]
	\caption{Performance of the different Teacher and Student Models on MNIST Test Set.}
	\label{table:mnist-metrics}
	\vskip -0.3in
	\begin{center}
		\begin{sc}
			\resizebox{\columnwidth}{!}{\begin{tabular}{lcccc}
					\toprule
					& \multicolumn{2}{c}{AlexNet Teacher} & \multicolumn{2}{c}{LeNet Student} \\
					Models & Accuracy & Loss & Accuracy  & Loss \\
					\midrule
					Baseline & 0.9958 & 0.020 & 0.9868 & \textbf{0.039} \\
					Mixup & 0.9960 & \textbf{0.010} & 0.9886 & 0.074 \\
					Cutmix & 0.9938 & 0.104 & 0.9861 & 0.107 \\
					\bottomrule
			\end{tabular}}
		\end{sc}
	\end{center}
\end{table}

\subsection{CIFAR 10 - Ablation Study - ResNet distilled to LeNet}

\subsubsection{Set Up}
To further explore the effect of data augmentation on distillation, we evaluated how augmentation impacted each step of the distillation process. This evaluation was done with ResNet18 teachers and LeNet students, with one of the outcomes (Augmented Teacher - Student) being the first experiment described above. We also evaluated the set ups of a Baseline ResNet teacher distilled into LeNet students trained with augmentation techniques. Finally, we evaluated augmented ResNet teachers distilled into augmented LeNet students. 

\begin{figure*}[htbp]
	\vskip 0.2in
	\begin{center}
		
		\begin{minipage}{0.33\textwidth}
			\centering
			\includegraphics[width=1.0\textwidth]{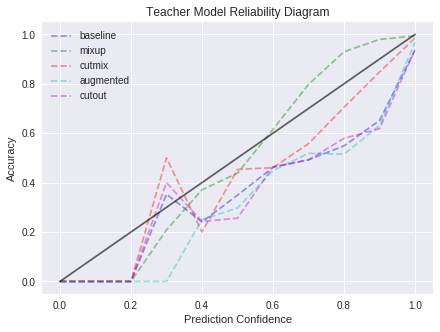}
			\subcaption{Teacher Model}
			
		\end{minipage}%
	\begin{minipage}{0.33\textwidth}
		\centering
		\includegraphics[width=1.0\textwidth]{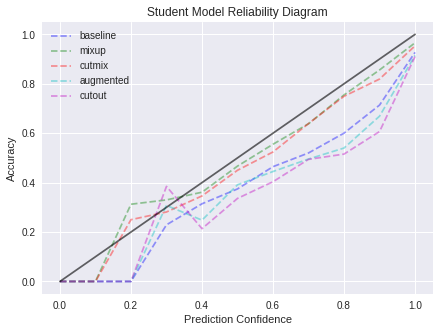}
		\subcaption{AlexNet Student Models}
	\end{minipage}
		\begin{minipage}{0.33\textwidth}
			\centering
			\includegraphics[width=1.0\textwidth]{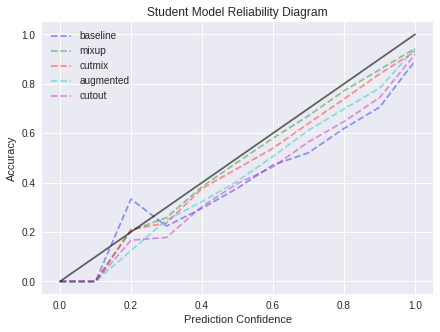}
			\subcaption{LeNet Student Models}
		\end{minipage}%
		
		\caption{Reliability curves for different models. It is observed that data augmentation generates calibrated predictions}
		\label{fig:reliability-image}
	\end{center}
	\vskip -0.2in
\end{figure*}

\subsubsection{Results}

This study showed that no matter the set up, data augmentation techniques hurt the performance of the student models. We suspect that much of this has to do with the interpolation of the latent space by the augmentation techniques. These results can be seen in \ref{table:student-metrics}, \ref{table:AugTeach-AugStudent}, \ref{table:baseTeach_AugStudent}. In our previous experiments, we found that the teacher manifold is the key driver in distillation, so it is no surprise that the techniques with augmented teachers performed poorly. However, it is interesting that adding augmentation to the student networks also inhibited distillation. However, more work may need to be done with more complex student networks to further understand the impact on augmentation in student networks.

\section{Explaining the Dichotomy}

It is clear from Table \ref{table:cifar-metrics} and from Figure  \ref{fig:overall} that when simpler models are distilled from cumbersome models trained with Data Augmentation, generalization performance is generally impaired. This is more pronounced for mixed sample data augmentation strategies like Mix Up and Cut Mix. This reversal of behavior is quite counter-intuitive, as we expect better generalized teacher models to transfer their capabilities to their respective students. We try to explain this in the following ways, with varying degrees of success.

\textbf{Latent Space Representations}: We visualize the latent representations generated under the influence of these different augmentation strategies. The teacher model representations as seen in \ref{fig:latent-space} depict a clearer picture that can help explain this reversal in performance. We plot the tSNE-reduced embeddings\cite{maaten2008visualizing} from the penultimate layer of the teacher ResNet models for four different classes in the CIFAR-10 Test Set. We chose two distinct pairs of semantically different classes. Shown in orange and green are two semantically similar classes, that represent vehicles(cars and trucks), whereas shown in blue and red are another pair of semantically similar classes (cat and bird), which are different from vehicles. This combination enables us to analyse relationships better, as a well-generalized model should be able to not only form correct clusters, but cluster similar classes near one another. Plotting all 10 classes might lead to undistinguishable inter-class distances, and so we focus on these representations. We note that for Cut Mix that uses a linear interpolation of both images and labels, the clusters are much more compressed and each semantic group lies much farther away from each other than any other strategy. Moreover, there is little interaction between the classes, as not a lot of points lie between the clusters. On the other hand, the baseline model manifold presents itself as much more uniform, wherein a lot of points lie on the boundaries of the class clusters and there is a gradual change in representational capability of the model between the classes. The two different semantic groups are closer in the baseline, when compared to any other augmentation strategy. This separability between classes and semantic groups exists in Cutout and transformation based Augmentations, but is not as pronounced as the interpolative Cut Mix. 

\textbf{Class-Separability}: To quantify the behavior observed in the latent space visualiztions, we propose the use of a very basic \textit{class-separability} metric that simply measures the difference in a model's confidence structure for different classes. In an ideal scenario where the model is supremely confident of each image, each class' confidence distribution is well separated and this can theoretically happen if a model is let to run for long enough using gradient descent algorithm. But, in real scenarios, models also assign probabilities to incorrect classes and as claimed by \cite{hinton2015distilling}, this relative probability structure is crucial in the generalization performance during distillation. The mathematical formulation of this \textit{class-separability} metric can be written as:

\begin{center}
	\[ S_f  = \frac{1}{C^2} \sum_{i=1}^{C} \sum_{j=1}^{C}  \mathcal{D_{KL}}(p_{f,i}, p_{f,j}) \]
\end{center}
Where, $p_{f,k} \in \mathrm{R}^C$ represents the average model prediction distribution for the class $k$, and $C$ represents the number of classes.  

 \begin{table}[htbp]
	\caption{Class Separability Score $S_f$}
	\label{table:separability}
	\begin{center}
		\begin{small}
		\begin{sc}
			\resizebox{\columnwidth}{!}{\begin{tabular}{lccc}
					\toprule
					
					Models & Teacher & LeNet & AlexNet \\
					\midrule
					Baseline & \textbf{3.73} & 3.21 & 3.07 \\
					Augment & 5.19 &  3.14 & 3.23 \\
					Cutout & 4.30 & 3.14 & 3.41  \\
					Mixup & 5.32 & 2.92 & \textbf{2.17}  \\
					Cutmix & 5.33 & \textbf{2.89} & 2.27 \\
					\bottomrule
			\end{tabular}}
		\end{sc}
		\end{small}
	\end{center}
\end{table}

\textbf{Class Discrimination}: Accuracy Performance can be a deceptive metric to evaluate generalization performance, as it depends on the one-hot representation of a prediction probability distribution. Similarly, the Similarity Score formulation we use above can be made even more robust by considering the latent space embeddings instead of the probability structure. We develop a novel metric, while taking inspiration from Discriminant Analysis techniques that help explain the model behavior in a better fashion with the current setup. This metric, called class-discrimination has been explained in Section 3.4 and involves a joint optimization of both intra-class similarity and inter-class similarity objectives in a latent space. We find the class-discrimination, intra-class similarity, and inter-class similarity metrics for the different models and find that data augmentation acts as a regularizer by creating more discriminant models. In simpler terms, a model with loose, yet well-separated(low intra-class similarity, low inter-class similarity) class representations will make a better teacher than a model with comparatively tighter and well-separated class representations(high intra-class similarity, low inter-class similarity), while not losing out on original accuracy performance.

 \begin{table}[htbp]
	\caption{Class Discrimination Evaluation for Teacher Models}
	\label{table:discrimination}
	\begin{center}
		\begin{sc}
			\resizebox{\columnwidth}{!}{\begin{tabular}{lllc}
					\toprule
					Metrics & Cohesion &  Adhesion & Discrimination \\
					\midrule
					Baseline & \textbf{0.739} & \textbf{-0.041} & \textbf{0.246} \\
					Augment & 0.886 &  -0.049 & 0.296 \\
					Cutout & 0.783 & -0.043 & 0.261  \\
					Mixup & 0.793 & -0.042 & 0.255  \\
					Cutmix & 0.917 & -0.050 & 0.306 \\
					\bottomrule
			\end{tabular}}
		\end{sc}
	\end{center}
\end{table}

As is evident from the above set of analysis, we can attempt to explain the adversarial impact of such augmentation strategies on distillation as that of the discriminative powers that these models gain during training. This is consistent with the ideas presented in \cite{he2019data}, wherein data augmentation strategies are believed to be regularization strategies that focus more on class-specific major features, while regularizing the example specific minor features. This regularization is more pronounced in interpolative techniques like Cut Mix and Mix Up as we believe they don't really add any new information to the model. By attempting to combine class-specific labels and corresponding images, they attempt to regularize on finer, example specific information and generate overall class-specific concepts as features. This is great for the model's performance on test data and data from a slightly different distribution, but distillation quality depends largely on the amount of information encoded in all the latent features encoded by the teacher model. Thus, if the generated features within any given class have greater variance, they are able to encode more information about the class' relationship with other classes and are expected to generate more generalized probability distributions. This is a key factor in generating better quality students. This also helps explaining the superior performance of Cutout and Augmentation that add new information by retaining the same label and transforming the image.

\begin{figure}[htbp]
	\vskip 0.2in
	\begin{center}
		
		\begin{minipage}{.25\textwidth}
			\centering
			\includegraphics[width=0.9\textwidth]{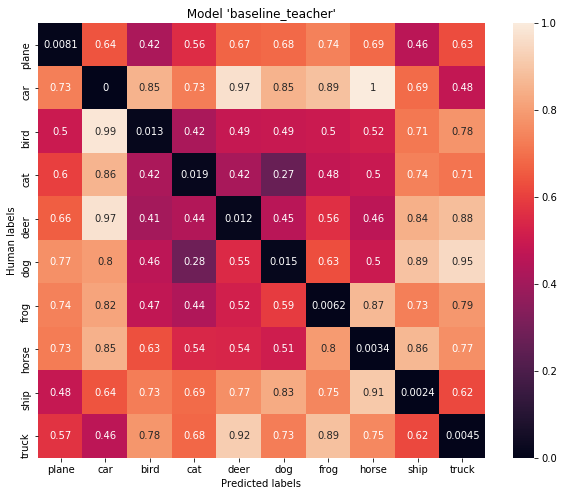}
			\subcaption{Baseline Teacher}
			
		\end{minipage}%
		\begin{minipage}{.25\textwidth}
			\centering
			\includegraphics[width=0.9\textwidth]{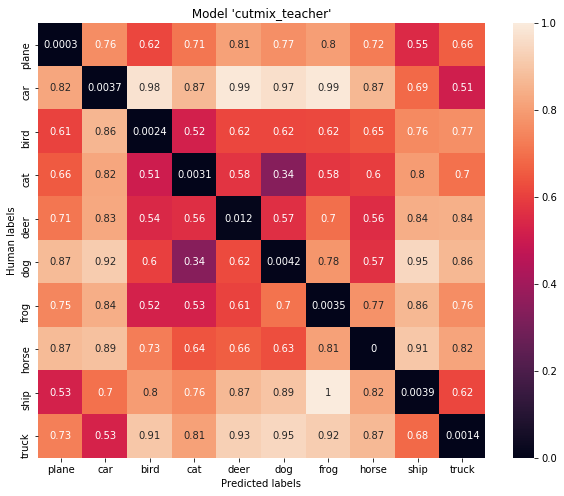}
			\subcaption{CutMix Teacher}
		\end{minipage}%

		\caption{Relative KL-Divergence Confusion Matrix}
		\label{fig:human-model-kld}
	\end{center}
	\vskip -0.3in
\end{figure}

\begin{figure}[h]
	\vskip 0.2in
	\begin{center}
		\centerline{\includegraphics[width=\columnwidth]{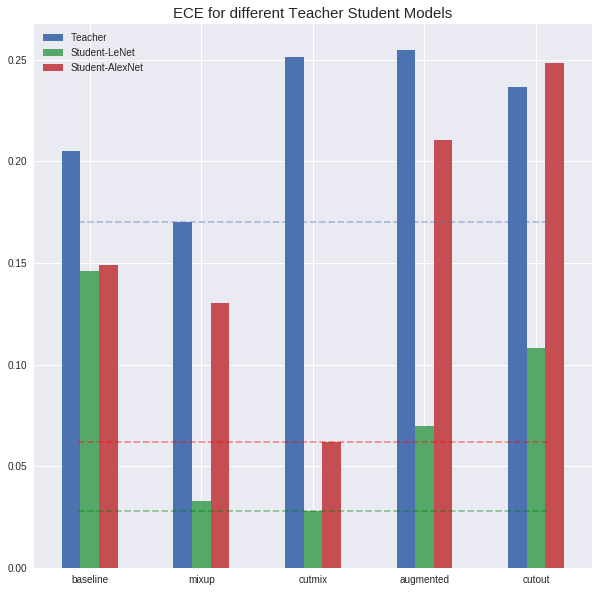}}
		\caption{Expected Calibration Error for the different Models. Smaller models like LeNet usually have better calibration performance and previous studies also show that linear interpolation of data samples always leads to better model calibration}
		\label{fig:ece-barplot}
	\end{center}
	\vskip -0.2in
\end{figure}

\textbf{Prediction Quality}: To test this idea of generating high quality probability distributions, we make use of the CIFAR-10 Human labeled datasets, and measure the quality of the average model confidence distributions against the human confidence estimates. This can again be represented as a KL-Divergence between the probability distributions of a model's average prediction probabilities for a given class against the average human distribution. We present this information in a Confusion Matrix like visualization in \ref{fig:human-model-kld}. Each cell represents the KL-Divergence between the human estimate on that class and the model's estimate. The KL-Divergences are all scaled uniformly and are color coded, so higher values correspond to brighter pixel values. We are more interested in the non-diagonal elements as they reveal the mutual information between classes and note that the Cut Mix Matrix is much brighter in those pixels when compared to the Baseline. This points out the fact that the mutual information between different classes is better encoded in the Baseline than Cut Mix-trained model. This enables the creation of a superior representation manifold for distillation to take place.

\textbf{Model Calibration}: We also measure calibration performance across the several models, and note that interpolative techniques generate better calibrated models consistently across teacher and student models. This is evident from the reliability diagrams in \ref{fig:reliability-image} and the Expected Calibration Error metrics in \ref{table:cifar-metrics} and \ref{fig:ece-barplot}. However, it has been hypothesized in \cite{guo2017calibration} that smaller models like LeNet generally tend to exhibit better Calibration performance than overparameterized, modern models like ResNet. This trend is visible as the reliability curves for all models tend to hug the ideal straight line closer as the model complexity decreases. However, no direct relationship can be found between the calibration performance of a student and the augmentation applied on its teacher model.

\textbf{Model Generalization}

Generalization is often defined in machine learning by the model's ability to adapt and classify on unseen data. However, with our analysis of the latent representation space, we found that as these techniques increase the intra-class discrimination, they may inhibit the ability to create probability distributions that mimic informative generalization for distillation into a student model. 

%

We analyzed this concept further through the use of confusion matrices. These confusion matrices compared the predicted label probabilities and actual classes. Both the baseline and Mix Up models show that they classify well overall, with a strong diagonal line in the un-masked confusion matrices. When that diagonal is masked,however, we can see that the probabilities between classes not chosen are far more spread apart in the baseline model. These can be compared to \ref{fig:human-dist}, which visualized the human generated predicted label probabilities. 

It is important to note that these matrices have different color scales, with the Masked Baseline going up to .1 and the Masked Cut Mix ranging to 0.07. The matrices show that they are both able to pick up on some of the nuances between data, but it can be seen that the masked baseline shows more generalization between classes with brighter colors between related classes. This ability to create an informative probabilistic distribution between different class types is the backbone for effective knowledge distillation.

\section{Future Work}

Moving ahead, a quantification of the regularization effect of mixed data augmentation strategies like Cut Mix and Mix Up should be formulated. An extension of this work would be to develop a novel augmentation strategy that is able to retain or improve the latent embedding manifold qualities for the given data set, and consequently, generate more generalized model predictions. A key missing aspect in this study is the fact that the training data set used to distill student models is the same as the one used to train the teacher models. A set of experiments could be conducted on a held out distillation data set. Another aspect of measuring true generalization performance could be the relative performance of these models under adversarial attack, and would be an interesting metric to analyse. 

%


\bibliography{example_paper}
\bibliographystyle{icml2019}





\appendix
\newpage
\newpage

\section{Appendix Figures}
\newpage

\newpage

\begin{table*}[htbp]
\caption{Appendix Table: All Student Metrics trained from ResNet18 CIFAR Teachers}
\label{table:student-metrics}
\centering
\begin{tabular}{@{}lllllll@{}}
\toprule
 & Model & Test Set & Accuracy & Precision & Recall & F1-Score \\ \midrule
\multirow{4}{*}{Baseline} & \multirow{2}{*}{LeNet} & CIFAR-10 & 0.652 & 0.648 & 0.652 & 0.650 \\
 &  & CINIC & 0.456 & 0.452 & 0.457 & 0.449 \\
 & \multirow{2}{*}{AlexNet} & CIFAR-10 & 0.768 & 0.767 & 0.769 & 0.768 \\
 &  & CINIC & 0.544 & 0.550 & 0.544 & 0.529 \\ \midrule
\multirow{4}{*}{Augmentation} & \multirow{2}{*}{LeNet} & CIFAR-10 & 0.631 & 0.632 & 0.631 & 0.474 \\
 &  & CINIC & 0.439 & 0.434 & 0.435 & 0.435 \\
 & \multirow{2}{*}{AlexNet} & CIFAR-10 & 0.777 & 0.776 & 0.777 & 0.766 \\
 &  & CINIC & 0.549 & 0.557 & 0.549 & 0.535 \\ \midrule
\multirow{4}{*}{Cut Out} & \multirow{2}{*}{LeNet} & CIFAR-10 & 0.644 & 0.643 & 0.644 & 0.643 \\
 &  & CINIC & 0.451 & 0.447 & 0.451 & 0.445 \\
 & \multirow{2}{*}{AlexNet} & CIFAR-10 & 0.785 & 0.783 & 0.785 & 0.784 \\
 &  & CINIC & 0.554 & 0.562 & 0.554 & 0.540 \\ \midrule
\multirow{4}{*}{Mix Up} & \multirow{2}{*}{LeNet} & CIFAR-10 & 0.633 & 0.631 & 0.633 & 0.632 \\
 &  & CINIC & 0.444 & 0.435 & 0.443 & 0.433 \\
 & \multirow{2}{*}{AlexNet} & CIFAR-10 & 0.714 & 0.712 & 0.714 & 0.712 \\
 &  & CINIC & 0.497 & 0.501 & 0.498 & 0.477 \\ \midrule
\multirow{4}{*}{Cut Mix} & \multirow{2}{*}{LeNet} & CIFAR-10 & 0.621 & 0.617 & 0.621 & 0.619 \\
 &  & CINIC & 0.439 & 0.434 & 0.439 & 0.435 \\
 & \multirow{2}{*}{AlexNet} & CIFAR-10 & 0.720 & 0.717 & 0.720 & 0.718 \\
 &  & CINIC & 0.498 & 0.502 & 0.498 & 0.478 \\ \bottomrule
\end{tabular}
\end{table*}

\newpage
\begin{table*}[htbp]
\caption{Appendix Table: All Baseline Teacher with Augmented Student Test Metrics trained from CIFAR-10}
\label{table:baseTeach_AugStudent}
\centering
\begin{tabular}{@{}lllllll@{}}
\toprule
 & Model & Test Set & Accuracy & Precision & Recall & F1-Score \\ \midrule
\multirow{2}{*}{Baseline} & \multirow{2}{*}{LeNet} & CIFAR-10 & 0.652 & 0.648 & 0.652 & 0.650\\
 &  & CINIC & 0.456 & 0.452 & 0.457 & 0.449 \\ \midrule
\multirow{2}{*}{Augmentation} & \multirow{2}{*}{LeNet} & CIFAR-10 & 0.641 & 0.639 & 0.641 & 0.639 \\
 &  & CINIC & 0.451 & 0.446 & 0.451 & 0.444 \\ \midrule
\multirow{2}{*}{Cut Out} & \multirow{2}{*}{LeNet} & CIFAR-10 & 0.635 & 0.630 & 0.635 & 0.632 \\
 &  & CINIC & 0.446 & 0.446 & 0.439 & 0.441  \\ \midrule
\multirow{2}{*}{Mix Up} & \multirow{2}{*}{LeNet} & CIFAR-10 & 0.647 & 0.645 & 0.647 & 0.647 \\
 &  & CINIC & 0.457 & 0.455 & 0.457 & 0.452 \\ \midrule
\multirow{2}{*}{Cut Mix} & \multirow{2}{*}{LeNet} & CIFAR-10 & 0.629 & 0.626 & 0.629 & 0.627 \\
 &  & CINIC & 0.439 & 0.433 & 0.439 & 0.434 \\ \bottomrule
\end{tabular}
\end{table*}

\begin{table*}[htbp]
\caption{Appendix Table: All ResNet Teacher - Student with same Augmentation Test Metrics trained from CIFAR-10}
\label{table:AugTeach-AugStudent}
\centering
\begin{tabular}{@{}lllllll@{}}
\toprule
 & Model & Test Set & Accuracy & Precision & Recall & F1-Score \\ \midrule
\multirow{2}{*}{Baseline} & \multirow{2}{*}{LeNet} & CIFAR-10 & 0.652 & 0.648 & 0.652 & 0.650\\
 &  & CINIC & 0.456 & 0.452 & 0.457 & 0.449 \\ \midrule
\multirow{2}{*}{Augmentation} & \multirow{2}{*}{LeNet} & CIFAR-10 & 0.641  & 0.637 & 0.641 & 0.638  \\
 &  & CINIC & 0.445  & 0.441 & 0.445 & 0.441 \\ \midrule
\multirow{2}{*}{Cut Out} & \multirow{2}{*}{LeNet} & CIFAR-10 & 0.636 & 0.643 & 0.636  & 0.638 \\
 &  & CINIC & 0.440 & 0.451 & 0.440  & 0.440  \\ \midrule
\multirow{2}{*}{Mix Up} & \multirow{2}{*}{LeNet} & CIFAR-10 & 0.626 & 0.622 & 0.626 & 0.624 \\
 &  & CINIC & 0.440  & 0.437  & 0.440  & 0.431  \\ \midrule
\multirow{2}{*}{Cut Mix} & \multirow{2}{*}{LeNet} & CIFAR-10 & 0.635 & 0.631 & 0.635 & 0.633 \\
 &  & CINIC & 0.446 & 0.443 & 0.446  & 0.441 \\ \bottomrule
\end{tabular}
\end{table*}


\begin{table*}[htbp]
\caption{Appendix Table: All Teacher Test Metrics trained from CIFAR-10}
\label{table:teacher-metrics}
\centering
\begin{tabular}{@{}lllllll@{}}
\toprule
 & Model & Test Set & Accuracy & Precision & Recall & F1-Score \\ \midrule
\multirow{2}{*}{Baseline} & \multirow{2}{*}{ResNet} & CIFAR-10 & 0.852 & 0.851 & 0.852 & 0.851 \\
 &  & CINIC & 0.600 & 0.602 & 0.600 & 0.593 \\ \midrule
\multirow{2}{*}{Augmentation} & \multirow{2}{*}{ResNet} & CIFAR-10 & 0.940 & 0.940 & 0.940 & 0.940 \\
 &  & CINIC & 0.630 & 0.620 & 0.628 & 0.629 \\ \midrule
\multirow{2}{*}{Cut Out} & \multirow{2}{*}{ResNet} & CIFAR-10 & 0.879 & 0.878 & 0.880 & 0.879 \\
 &  & CINIC & 0.629 & 0.620 & 0.628 & 0.630 \\ \midrule
\multirow{2}{*}{Mix Up} & \multirow{2}{*}{ResNet} & CIFAR-10 & 0.867 & 0.954 & 0.867 & 0.953 \\
 &  & CINIC & 0.603 & 0.606 & 0.603 & 0.598 \\ \midrule
\multirow{2}{*}{Cut Mix} & \multirow{2}{*}{ResNet} & CIFAR-10 & 0.954 & 0.869 & 0.954 & 0.868 \\
 &  & CINIC & 0.710 & 0.707 & 0.720 & 0.710 \\ \bottomrule
\end{tabular}
\end{table*}

\newpage

\end{document}